\documentclass[11pt]{article}
\usepackage{dialogue2021}

\usepackage{ifxetex}
\ifxetex
    \usepackage{fontspec}
    \setromanfont{Times New Roman}
\else
  \usepackage[T1]{fontenc}
  \usepackage[utf8]{inputenc}
  \usepackage{cmap}
  \usepackage{times}
  \usepackage{latexsym}

\fi

\usepackage[russian,british]{babel}
\usepackage{url}
\usepackage{pgf}

\usepackage{covington} 

\dialogfinalcopy 

\title{Transformers for Headline Selection for Russian News Clusters}

\author{Pavel Voropaev \\
  Moscow Institute\\of Physics and Technology \\
  Moscow, Russia \\
  {\tt voropaev@phystech.edu} \\\And
  Olga Sopilnyak \\
  Moscow Institute\\of Physics and Technology \\
  Moscow, Russia \\
  {\tt olga.sopilnyak@phystech.edu} \\}

\date{}

\begin{document}
\maketitle
\begin{abstract}
  In this paper, we explore various multilingual and Russian pre-trained transformer-based models for the Dialogue Evaluation 2021 shared task on headline selection. Our experiments show that the combined approach is superior to individual multilingual and monolingual models. We present an analysis of a number of ways to obtain sentence embeddings and learn a ranking model on top of them. We achieve the result of 87.28\% and 86.60\% accuracy for the public and private test sets respectively.
  
  
\end{abstract}



  

\section{Introduction}
\label{intro}

%
%

News stories clustering task not only has a wide application area in the industry, but also helps to explore the usage boundaries of sentence embeddings obtained with different models. For example, news aggregators actively use clustering algorithms to generate news feeds from different sources and to select a single headline. The recent progress in designing multilingual models \cite{hu2020xtreme}, trained for dozens or even hundreds of languages at once, makes it possible to use them for monolingual tasks, particularly for Russian language tasks \cite{shavrina2020russiansuperglue}. At the same time, Russian BERT-based models are actively evolving, and their comparison with more universal multilingual ones may be of interest.

The task of generating or selecting headlines for a single news cluster has a wide range of applications along with clustering. However, even considering the current level of generative models progress, the presence of great interest to them and strong state-of-the-art models \cite{brown2020language}, it is not always possible to use this models in the industry due to the unwarranted quality of generated texts and high demand for computational resources. An alternative choice is to select ready-made headings among those presented in the cluster. This task can be solved as a classification or ranking problem.

In this paper, the cluster headline selection task is considered. We did not find much related work except \cite{selectinglabels} where a simple rule-based system is proposed. We use the corpora provided by the Dialogue Evaluation 2021 shared task organizers \cite{news_clustering} and introduce the solution that has shown the best results among other participants. Our code is publicly available at \url{https://github.com/sopilnyak/headline-selection}.

\section{Experimental evaluation}

The training corpus for choosing the best headline is proposed at the Dialogue Evaluation 2021 shared task. It consists of pairs of news identifiers (URLs) each one corresponding to one of four tags: \texttt{left}, \texttt{right}, \texttt{draw}, or \texttt{bad}. The last label means that the authors of the markup have identified the pair as a clustering error. The test set is divided into two parts: for public and private leaderboard, each containing news headlines for two specific dates: May 27, 2020 and May 29, 2020. To evaluate the result, a weighted accuracy is used, while the \texttt{bad} label is omitted, and the weights for the remaining labels are shown in Table \ref{accuracy_weights}.

\begin{table}[t]
\begin{center}
\begin{tabular}{l|ccc}
\hline
 & \bf left & \bf right & \bf draw \\\hline
\bf left & 1 & 0 & 0.5 \\\hline
\bf right & 0 & 1 & 0.5 \\\hline
\bf draw & 0.5 & 0.5 & 1 \\
\hline
\end{tabular}
\end{center}
\caption{Accuracy weights for headline selection task evaluation.}
\label{accuracy_weights} 
\end{table}

\subsection{Embeddings}

For each headline from the training corpus, embeddings are obtained from various Russian and multilingual models. We use pretrained BERT-based models trained both on Russian monolingual corpora (RuBERT \cite{DBLP:journals/corr/abs-1905-07213}, SBERT \cite{DBLP:journals/corr/abs-1908-10084}) and in multiple languages (mT5 \cite{xue2021mt5}, XLM-RoBERTa \cite{DBLP:journals/corr/abs-1911-02116}) including Russian. In addition, a multilingual version of USE \cite{DBLP:journals/corr/abs-1803-11175} embeddings is used. The mentioned models show state-of-the-art results on a number of NLP benchmarks \cite{hu2020xtreme}, including those in Russian language \cite{shavrina2020russiansuperglue}, so it was natural to test them on the task of selecting the best headline for the cluster.

To obtain a headline embedding, we take the average word embeddings from the layer 19 (of 25) for SBERT, XLM-R and mT5 and layer 8 (of 13) for RuBERT, considering the length of the headline. In the case of mT5 model, which is trained mainly for solving seq2seq tasks, the decoder is removed and the embedding is taken from layer 19 of the encoder. We use recommended pretrained tokenizers from the \texttt{transformers} library \cite{wolf-etal-2020-transformers}. These tokenizers are based on WordPiece and SentencePiece models \cite{DBLP:journals/corr/abs-1808-06226}.

\subsection{Classification}

Then we train a classifier on top of the embeddings. We use CatBoost ranking model \cite{DBLP:journals/corr/abs-1810-11363}, which is a gradient boosting over decision trees algorithm. Pairs of headline embeddings are fed to the classifier as input, while the best headline in the pair is considered the ''positive'' element of the pair, and the other one is considered ''negative''. We chose \texttt{PairLogitLoss} (1) as the target loss function.

\begin{equation}
\texttt{PairLogitLoss}(a) = -\sum\limits_{p, n ~\in~ \texttt{Pairs}} \log\left(\frac{1}{1 + e^{-(a_p - a_n)}}\right)
\end{equation}

An ensemble of ranking models is trained based on different features. The number of decision trees in CatBoost is set to $10^3$, the best epoch is chosen based on the validation score. We obtain the final headline rank by averaging the ranks predicted by each of the models, and then each pair is assigned one of the \texttt{left}, \texttt{right}, or \texttt{draw} labels depending on the resulting rank difference $r_r - r_l$. More specifically, the rank difference $r_r - r_l < 0$ means the winner is \texttt{left}, $r_r - r_l > 0$ means the winner is \texttt{right}, and $|r_r - r_l| \leq 0.1$ correspond to \texttt{draw}.

\begin{table}[t]
\begin{center}
\begin{tabular}{l|ccc}
\hline
\bf Model & \bf Validation & \bf Public LB test & \bf Private LB test \\\hline
SBERT & 85.98 & 84.48 & 83.41 \\
RuBERT & 83.97 & 81.38 & 81.64 \\
XLM-R & 87.93 & 84.30 & 84.13 \\
mT5 & 88.52 & 84.48 & 82.60 \\
USE & 81.23 & 80.79 & 80.68 \\
Blend-5 & \textbf{88.77} & \textbf{87.28} & \textbf{86.60} \\
\hline
\end{tabular}
\end{center}
\caption{We report the accuracy on custom validation set and two test sets. The best results for each set are in bold. All the results are averaged over five different training runs. The ensemble of five models (Blend-5) achieves the result of 87.28\% and 86.60\% accuracy for the public and private test sets respectively.}
\label{headline_results} 
\end{table}

To explore models and compare the results, we train classifiers on top of each type of embeddings separately. Every model was trained on an Nvidia Tesla P100 GPU provided by Google Colab. Table \ref{headline_results} shows the results of using various embeddings and training classifiers on top of them. The table shows that the best result is obtained by the ensemble of the five models mentioned in the paper (referred as Blend-5), but the single multilingual model mT5 shows a comparable accuracy.

During the Dialogue Evaluation 2021 competition we achieved 86.00\% and 85.40\% accuracy for the public and private test sets by taking the average word embeddings in the top layer. But further experiments show that middle layers perform better than top layers with the result of 87.28\% and 86.60\% accuracy respectively.

\subsection{Analysis and results}

In this section, we analyze the model and explore the impact of several aspects of out approach. We list some examples of wrongly predicted labels and report the evaluation results for different variants of the model on the private LB test set.

\textbf{Error analysis.} We selected 300 examples where the model confused between \texttt{left} and \texttt{right} labels and skipped examples with \texttt{draw} label. Errors can be divided into several types as listed at (\ref{ex1})--(\ref{ex4}) together with several corresponding examples and their translation to English. The model is more likely to prefer titles with lack of facts and sometimes tends to choose verbose headlines. Other wrongly selected headlines include opinions and biased titles. Finally, having a pair of almost equivalent titles, the model could choose the wrong label.

\begin{examples}

\item \textbf{Headlines containing insufficient facts} \\
\label{ex1}

\textbf{gold: }{\selectlanguage{russian}Ту-22МЗМ Казанского авиазавода испытали на сверхзвуке} \\
\phantom{\textbf{gold: }}{\it Kazan aircraft factory's Tu-22MZM tested at supersonic mode}

\textbf{pred: }{\selectlanguage{russian}В ОПК рассказали об испытаниях модернизированного ракетоносца Ту-22М3М} \\
\phantom{\textbf{pred: }}{\it The defense industry told about the tests of the modernized Tu-22M3M missile carrier}

\textbf{gold: }{\selectlanguage{russian}День проведения парада Победы будет нерабочим --- Песков} \\
\phantom{\textbf{gold: }}{\it Victory Parade day will be non-working --- Peskov}

\textbf{pred: }{\selectlanguage{russian}Песков заявил о большой вероятности объявления еще одного выходного} \\
\phantom{\textbf{pred: }}{\it Peskov said about the high probability of announcing another weekend}

\hspace{0.5em}

\item \textbf{Verbose headlines} \\

\textbf{gold: }{\selectlanguage{russian}С 27 мая москвичи могут бесплатно сдать тест на антитела} \\
\phantom{\textbf{gold: }}{\it From May 27, Moscow residents can take an antibody test for free}

\textbf{pred: }{\selectlanguage{russian}<<Как и где сдать тест на антитела к коронавирусу в Москве. С 27 мая это может сделать любой желающий} \\
\phantom{\textbf{pred: }}{\it How and where to take an antibody test against coronavirus in Moscow. Anyone can do it from May 27}

\textbf{gold: }{\selectlanguage{russian}Украинский суд признал нацистской символику дивизии СС <<Галичина>>} \\
\phantom{\textbf{gold: }}{\it The Ukrainian court found that the symbols of the SS division ''Galicia'' were Nazi}

\textbf{pred: }{\selectlanguage{russian}<<Победа справедливости, здравого смысла и закона>>: Вятрович проиграл суд по делу о символике СС <<Галичина>>} \\
\phantom{\textbf{pred: }}{\it ''Victory of justice, common sense and law'': Vyatrovich lost the court case concerning the SS ''Galicia'' symbols}

\item \textbf{Biased headlines} \\

\textbf{gold: }{\selectlanguage{russian}Роскомнадзор начнет блокировать в России пиратские приложения} \\
\phantom{\textbf{gold: }}{\it Roskomnadzor will begin to block pirated applications in Russia}

\textbf{pred: }{\selectlanguage{russian}Госдума приняла спорный законопроект о блокировке приложений с пиратским контентом} \\
\phantom{\textbf{pred: }}{\it State Duma passed controversial bill on blocking applications with pirated content}

\textbf{gold: }{\selectlanguage{russian}Доллар в обменниках ускорил рост} \\
\phantom{\textbf{gold: }}{\it Dollar in exchange offices accelerated growth}

\textbf{pred: }{\selectlanguage{russian}Заманивают иностранцев под покупку облигаций. Почему гривня снова падает} \\
\phantom{\textbf{pred: }}{\it Luring foreigners to buy bonds. Why is the hryvnia falling again}

\item \textbf{Equivalent pairs of headlines} \\
\label{ex4}

\textbf{gold: }{\selectlanguage{russian}Россияне массово забирают валюту из банков} \\
\phantom{\textbf{gold: }}{\it Russians massively withdraw currency from banks}

\textbf{pred: }{\selectlanguage{russian}Жители страны массово снимают валюту с банковских счетов} \\
\phantom{\textbf{pred: }}{\it Residents of the country are massively withdrawing currency from bank accounts}

\textbf{gold: }{\selectlanguage{russian}Россиянам разъяснили, когда можно будет поехать в отпуск за рубеж} \\
\phantom{\textbf{gold: }}{\it Russians were clarified when it will be possible to go on vacation abroad}

\textbf{pred: }{\selectlanguage{russian}Россиянам озвучили возможные сроки возобновления поездок за границу} \\
\phantom{\textbf{pred: }}{\it Russians were announced the possible date of the resumption of trips abroad}

\end{examples}

Thus, a good headline can be defined as precise, short, unbiased, and containing as many significant facts as possible. However, classifiers based on language models can rank headlines which do not meet these criteria higher than manually selected ones. We assume that adding more training data or introducing multitask learning, combining another training objectives, such as information extraction, could help achieve better results.

\textbf{Sentence representations.} We explore a different way to obtain sentence embeddings from language model's top layer output: using the embeddings of the first token, known as [CLS] token, which is sometimes more common than averaging the word embeddings. We compare the results for all models, except mT5, which doesn't have the [CLS] token in its dictionary. However, as shown in Table \ref{ablation}, averaging show slightly better results for most of the models. We assume this is because embeddings of the [CLS] token contain high-level semantic meaning \cite{hu2020xtreme}, while for headline selection task it is more important not to lose the token-level information to meet the previously formulated criteria of a good headline. Moreover, we analyze whether sentence embeddings in the middle layers can be more suitable than in the last ones. Experiments show that layers 17 to 19 (of 25) for SBERT, XLM-R and mT5 and layers 8 to 9 (of 13) for RuBERT perform better than the top layers. The reason for this may be the same: top layers embed more high-level semantic meaning than the middle ones.

\begin{table}[t]
\begin{center}
\begin{tabular}{l|ccccc}
\hline
\bf Sentence representation & \bf SBERT & \bf RuBERT & \bf XLM-R & \bf mT5 & \bf Blend-5 \\\hline
Top layer: average embeddings & 78.62 & 75.50 & 80.01 & 81.60 & 85.20 \\
Top layer: [CLS] token embeddings & 77.77 & 76.53 & 75.48 & --- & 84.50 \\
Layer 23/11: average embeddings & 81.23 & 78.65 & 82.59 & 81.83 & 85.64 \\
Layer 21/9: average embeddings & 83.31 & 81.41 & 83.30 & 82.48 & 86.31 \\
Layer 19/8: average embeddings & 83.41 & \bf 81.64 & \bf 84.13 & 82.60 & 86.60 \\
Layer 17/8: average embeddings & \bf 83.55 & \bf 81.64 & 84.06 & \bf 83.56 & \bf 86.93 \\
\hline
\end{tabular}
\end{center}
\caption{Comparison of sentence representations. The table shows accuracy on private LB test set. Sentence representations are either averaged token embeddings in middle and top layers, or [CLS] token embeddings in the top layer. Layer N/M means that for SBERT, XLM-R and mT5 we take layer N and for RuBERT we take layer M.}
\label{ablation} 
\end{table}

\section{Conclusion}

In this paper, we study applications of pre-trained models for headline selection and demonstrate the superiority of ensembles of modern BERT-based models. We have shown that multilingual models, such as mT5, demonstrate decent results in the task, and are superior to the single-language models in the same conditions.

Further research can be made in additional training of the top layers of multilingual models on Russian-language corpora, as well as more fine-tuning of lightweight models, such as multilingual USE or LASER \cite{DBLP:journals/corr/abs-1812-10464}, to reduce system requirements for headline selection.

\bibliography{dialogue.bib}

\begin{thebibliography}{10}
\def\selectlanguageifdefined#1{
\expandafter\ifx\csname date#1\endcsname\relax
\else\selectlanguage{#1}\fi}
\providecommand*{\href}[2]{{\small #2}}
\providecommand*{\url}[1]{{\small #1}}
\providecommand*{\BibUrl}[1]{\url{#1}}
\providecommand{\BibAnnote}[1]{}
\providecommand*{\BibEmph}[1]{#1}
\ProvideTextCommandDefault{\cyrdash}{\iflanguage{russian}{\hbox
  to.8em{--\hss--}}{\textemdash}}
\providecommand*{\BibDash}{\ifdim\lastskip>0pt\unskip\nobreak\hskip.2em plus
  0.1em\fi
\cyrdash\hskip.2em plus 0.1em\ignorespaces}
\renewcommand{\newblock}{\ignorespaces}

\bibitem{DBLP:journals/corr/abs-1812-10464}
\selectlanguageifdefined{english}
\BibEmph{Artetxe~Mikel, Schwenk~Holger}. Massively Multilingual Sentence
  Embeddings for Zero-Shot Cross-Lingual Transfer and Beyond~// \BibEmph{CoRR}.
  \BibDash
\newblock 2018. \BibDash
\newblock Vol. abs/1812.10464. \BibDash
\newblock \href{http://arxiv.org/abs/1812.10464}{1812.10464}.

\bibitem{DBLP:journals/corr/abs-1810-11363}
\selectlanguageifdefined{english}
\BibEmph{Dorogush~Anna~Veronika, Ershov~Vasily, Gulin~Andrey}. CatBoost:
  gradient boosting with categorical features support~// \BibEmph{CoRR}.
  \BibDash
\newblock 2018. \BibDash
\newblock Vol. abs/1810.11363. \BibDash
\newblock \href{http://arxiv.org/abs/1810.11363}{1810.11363}.

\bibitem{news_clustering}
\selectlanguageifdefined{english}
\BibEmph{Gusev~Ilya;~Smurov~Ivan}. Russian News Clustering and Headline
  Selection Shared Task~// Computational Linguistics and Intellectual
  Technologies: Papers from the Annual Conference ``Dialogue''. \BibDash
\newblock 2021.

\bibitem{DBLP:journals/corr/abs-1808-06226}
\selectlanguageifdefined{english}
\BibEmph{Kudo~Taku, Richardson~John}. SentencePiece: {A} simple and language
  independent subword tokenizer and detokenizer for Neural Text Processing~//
  \BibEmph{CoRR}. \BibDash
\newblock 2018. \BibDash
\newblock Vol. abs/1808.06226. \BibDash
\newblock \href{http://arxiv.org/abs/1808.06226}{1808.06226}.

\bibitem{DBLP:journals/corr/abs-1905-07213}
\selectlanguageifdefined{english}
\BibEmph{Kuratov~Yuri, Arkhipov~Mikhail}. Adaptation of Deep Bidirectional
  Multilingual Transformers for Russian Language~// \BibEmph{CoRR}. \BibDash
\newblock 2019. \BibDash
\newblock Vol. abs/1905.07213. \BibDash
\newblock \href{http://arxiv.org/abs/1905.07213}{1905.07213}.

\bibitem{brown2020language}
\selectlanguageifdefined{english}
\BibEmph{Brown~Tom~B., Mann~Benjamin, Ryder~Nick et~al.} Language Models are
  Few-Shot Learners. \BibDash
\newblock 2020. \BibDash
\newblock 2005.14165.

\bibitem{DBLP:journals/corr/abs-1908-10084}
\selectlanguageifdefined{english}
\BibEmph{Reimers~Nils, Gurevych~Iryna}. Sentence-BERT: Sentence Embeddings
  using Siamese BERT-Networks~// \BibEmph{CoRR}. \BibDash
\newblock 2019. \BibDash
\newblock Vol. abs/1908.10084. \BibDash
\newblock \href{http://arxiv.org/abs/1908.10084}{1908.10084}.

\bibitem{shavrina2020russiansuperglue}
\selectlanguageifdefined{english}
RussianSuperGLUE: A Russian Language Understanding Evaluation Benchmark~/
  Tatiana~Shavrina, Alena~Fenogenova, Anton~Emelyanov et~al.~// \BibEmph{arXiv
  preprint arXiv:2010.15925}. \BibDash
\newblock 2020.

\bibitem{selectinglabels}
\selectlanguageifdefined{english}
\BibEmph{Thirunarayan~Krishnaprasad, Immaneni~Trivikram, Shaik~Mastan}.
  \href{http://dx.doi.org/10.1007/978-3-540-73351-5_11}{Selecting Labels for
  News Document Clusters}. \BibDash
\newblock 2007. \BibDash 01. \BibDash
\newblock P.~119--130.

\bibitem{wolf-etal-2020-transformers}
\selectlanguageifdefined{english}
Transformers: State-of-the-Art Natural Language Processing~/ Thomas~Wolf,
  Lysandre~Debut, Victor~Sanh et~al.~// Proceedings of the 2020 Conference on
  Empirical Methods in Natural Language Processing: System Demonstrations.
  \BibDash
\newblock Online~: Association for Computational Linguistics, 2020. \BibDash
  Oct. \BibDash
\newblock P.~38--45. \BibDash
\newblock Access mode:
  \BibUrl{https://www.aclweb.org/anthology/2020.emnlp-demos.6}.

\bibitem{DBLP:journals/corr/abs-1803-11175}
\selectlanguageifdefined{english}
Universal Sentence Encoder~/ Daniel~Cer, Yinfei~Yang, Sheng{-}yi~Kong et~al.~//
  \BibEmph{CoRR}. \BibDash
\newblock 2018. \BibDash
\newblock Vol. abs/1803.11175. \BibDash
\newblock \href{http://arxiv.org/abs/1803.11175}{1803.11175}.

\bibitem{DBLP:journals/corr/abs-1911-02116}
\selectlanguageifdefined{english}
Unsupervised Cross-lingual Representation Learning at Scale~/ Alexis~Conneau,
  Kartikay~Khandelwal, Naman~Goyal et~al.~// \BibEmph{CoRR}. \BibDash
\newblock 2019. \BibDash
\newblock Vol. abs/1911.02116. \BibDash
\newblock \href{http://arxiv.org/abs/1911.02116}{1911.02116}.

\bibitem{hu2020xtreme}
\selectlanguageifdefined{english}
XTREME: A Massively Multilingual Multi-task Benchmark for Evaluating
  Cross-lingual Generalization~/ Junjie~Hu, Sebastian~Ruder, Aditya~Siddhant
  et~al. \BibDash
\newblock 2020. \BibDash
\newblock 2003.11080.

\bibitem{xue2021mt5}
\selectlanguageifdefined{english}
\BibEmph{Xue~Linting, Constant~Noah, Roberts~Adam et~al.} mT5: A massively
  multilingual pre-trained text-to-text transformer. \BibDash
\newblock 2021. \BibDash
\newblock 2010.11934.

\end{thebibliography}
\bibliographystyle{ugost2008ls}



\end{document}